\documentclass[letterpaper, 10 pt, conference]{ieeeconf}  % Comment this line out if you need a4paper

\IEEEoverridecommandlockouts
\usepackage[space, compress, sort]{cite}

\usepackage{graphics} % for pdf, bitmapped graphics files
\usepackage{epsfig} % for postscript graphics files
\usepackage{mathptmx} % assumes new font selection scheme installed
\usepackage{times} % assumes new font selection scheme installed
\usepackage{amsmath} % assumes amsmath package installed
\usepackage{todonotes}
\usepackage{siunitx}
\usepackage{amssymb, mathtools}  % assumes amsmath package installed
\makeatletter
\def\@opargbegintheorem#1#2#3{\trivlist
   \item[]{\bfseries #1\ #2\ (#3)} \itshape}
\makeatother

\usepackage{float}
\usepackage[breaklinks=true,bookmarks=true,colorlinks]{hyperref}
\usepackage{graphicx}
\usepackage{booktabs}   % for \toprule, \midrule, \
\usepackage{multirow}   % for \multirow cells
\usepackage{xcolor}
\usepackage{multirow}  % For merging rows
\usepackage{adjustbox}
\usepackage{float}
\usepackage[font=small]{caption}
\usepackage{algorithm}
\usepackage{algorithmic}
\usepackage{xspace}

\hypersetup{hidelinks}

\newif\ifsubmission

\submissionfalse       % ARXIV VERSION

\title{\LARGE \bf Data-Efficient Adaptation of a Driving VLA to Class 8 Trucks}

\ifsubmission

\author{Anonymous Authors}

\else

\author{
Satyajeet Das$^{1}$,
Aaron Buxbaum$^{2}$,
Niels Joubert$^{2}$,
and Gaurav S. Sukhatme$^{1}$%
\thanks{$^{1}$Satyajeet Das and Gaurav S. Sukhatme are with the
Department of Computer Science, University of Southern California,
Los Angeles, CA 90089, USA.
Emails: \{satyajee, gaurav\}@usc.edu.}%
\thanks{$^{2}$Aaron Buxbaum and Niels Joubert are with Stack AV,
Pittsburgh, PA, USA.
Emails: \{abuxbaum, njoubert\}@stackav.com.}%
}
\fi

\begin{document}

\maketitle
\thispagestyle{empty}
\pagestyle{empty}

%%%%%%%%%%%%%%%%%%%%%%%%%%%%%%%%%%%%%%%%%%%%%%%%%%%%%%%%%%%%%%%%%%%%%%%%%%%%%%%%

% =========================================================
% ABSTRACT
% =========================================================
\begin{abstract}

\ifsubmission
\else
% =========================================================
% ARXIV ABSTRACT
% =========================================================
Class~8 trucks differ from passenger cars in geometry, dynamics, and maneuvering requirements. As a result, vision-language-action (VLA) models trained for passenger vehicles do not readily transfer to Class 8 trucks, particularly in unstructured scenarios such as accident scenes and construction zones. Rather than training a truck-driving VLA from scratch, we propose an \emph{adapt-then-steer} strategy that adapts an off-the-shelf VLA to generate trajectories for Class-8 trucks in these challenging scenarios. 
In the \emph{adapt} stage, we use NVIDIA's \textit{Alpamayo~1.5} as the base model, fine-tuning only its action-generation stack on a few hundred real-world construction and accident-related highway scenarios. In the \emph{steer} stage, we introduce \emph{Flow Velocity Steering} (FVS) to further refine the model's predictions while holding the adapted VLA fixed.
FVS is a compact, flow-time-conditioned residual module that adds learned corrections to the action-space flow velocity used to update the action sequence at each generation step. In open-loop evaluation on a scenario-disjoint held-out set, targeted fine-tuning more than halves single-candidate average displacement error (ADE) and final displacement error (FDE) over the entire $6.4\,\mathrm{s}$ horizon compared to the base model.
Using the same targeted demonstrations, FVS further reduces the fine-tuned model's full-horizon ADE and FDE by \(13.9\%\) and \(16.5\%\), respectively.
At matched data budgets, targeted supervision yields \(19\)-\(26\%\) lower full-horizon ADE than general truck-driving supervision, while the targeted model remains competitive with a model fine-tuned on approximately \(65\times\) as many general truck-driving scenarios.
These results support \emph{adapt-then-steer} for data-efficient vehicle-domain transfer to Class~8 trucks. Our project website is available at
\href{https://truckvla.github.io/}{https://truckvla.github.io}.

\fi

\end{abstract}
\section{Introduction}
\label{sec:introduction}

\begin{figure}[t]
    \centering
    \includegraphics[width=\columnwidth]{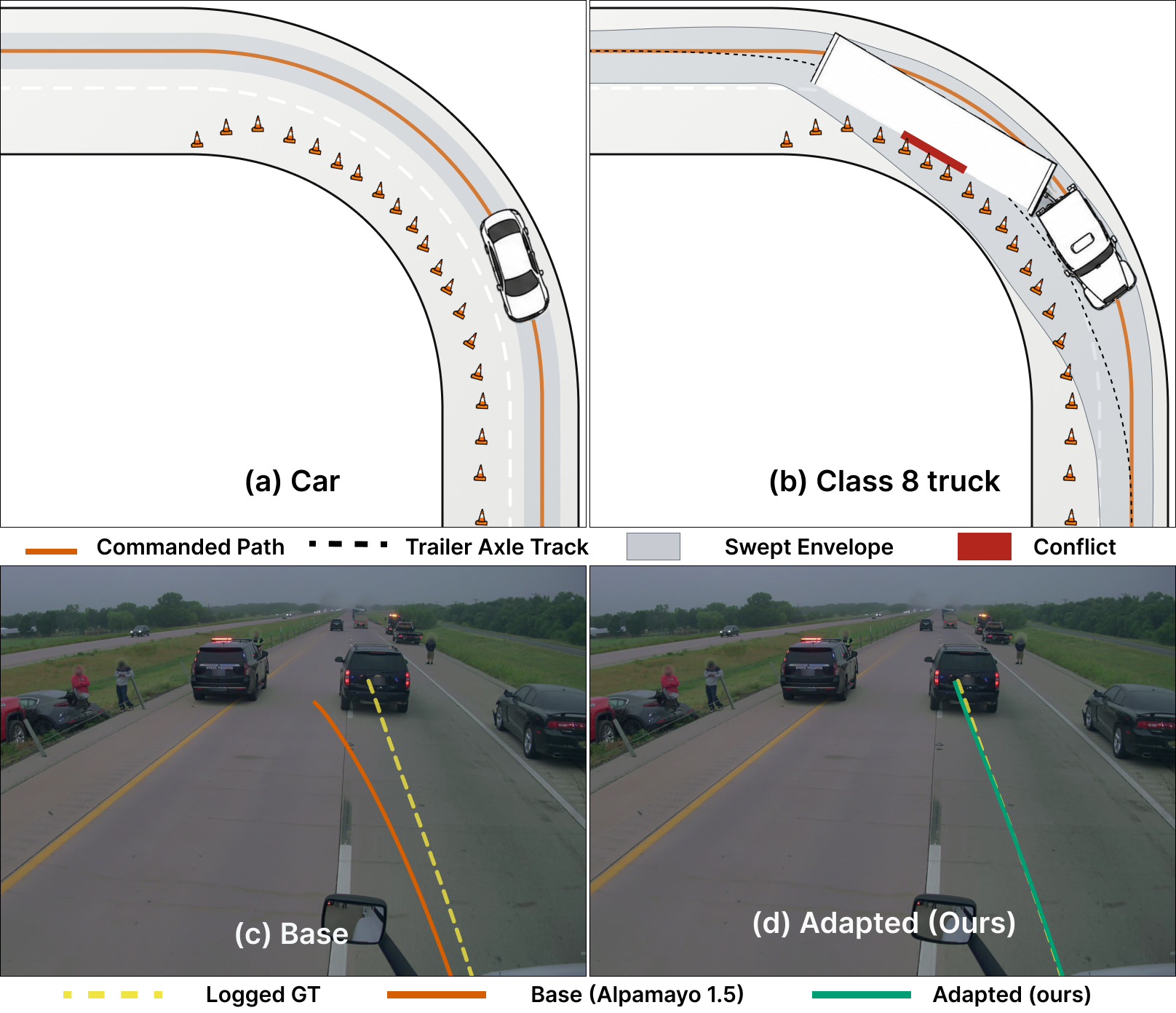}
   \caption{
\textbf{Motivation and qualitative example of truck-domain adaptation.}
\textbf{(a)-(b)} 
% Under the same illustrative reference path, the
% passenger car clears the work-zone closure, whereas trailer
% off-tracking causes the Class~8 tractor-trailer's swept envelope to
% enter it.
Under the same reference path, a
passenger car clears the work-zone closure, whereas trailer
off-tracking causes the Class~8 truck's swept envelope
to enter the closure.
\textbf{(c)-(d)} In a held-out accident-related event, off-the-shelf
Alpamayo~1.5 deviates from the logged expert truck trajectory, whereas
the adapted model follows it more closely.
\emph{Adapted} denotes Targeted Truck-SFT followed by FVS.
All displayed trajectories in (c)-(d) span \(6.4\,\mathrm{s}\).
The schematics in (a)-(b) are illustrative and not to scale.
}
\vspace{-1em}

\label{fig:motivation}
\end{figure}

Vision-language-action (VLA) models for autonomous driving combine multimodal scene representations with trajectory generation, offering a promising starting point for handling complex driving scenarios~\cite{hwang2024emma,nvidia2025alpamayo}. However, the trajectory appropriate for a scene depends on the vehicle's physical embodiment.
A Class~8 truck 
% (hereafter, \emph{truck}) 
differs from a passenger car in sensor viewpoint, articulated geometry, kinematics, vehicle dynamics, and maneuver timing~\cite{fent2024truckscenes,ghilotti2026truckdrive,oh2026mvadapt}. Figure~\ref{fig:motivation}(a)-(b) illustrates one such geometric
consequence: under the same reference path, the passenger car clears
a work zone, whereas trailer off-tracking causes the truck's
swept envelope to enter it. More broadly, one vehicle's trajectory may not satisfy the geometric and maneuverability requirements of another.

% These constraints become especially critical in construction zones and accident-related highway disruptions, which we refer to as the \emph{long-tail scenarios}. 
% These constraints become especially critical in construction zones and accident-related highway disruptions, therefore we choose to study the two target long-tail scenario families studied in this paper.
These constraints become especially critical in construction zones and accident-related highway disruptions, the two target long-tail scenario families studied in this paper.
% Temporary lane closures or obstructions in these settings frequently require departures from ordinary lane following that must accommodate the truck's dynamics. 
Temporary lane closures and
obstructions in these settings can require trajectories that depart
from nominal lane following while respecting truck-specific
geometric and dynamic constraints.
To study transfer to this setting, we use NVIDIA's Alpamayo~1.5~\cite{nvidia2025alpamayo,nvidia2026alpamayo15}, an open-weight driving VLA that generates continuous trajectories from multimodal driving observations. We evaluate it as an open-loop candidate-trajectory generator intended to provide proposals for downstream planning and safety validation, or to support remote-assist personnel. However, its off-the-shelf predictions deviate substantially from logged expert truck trajectories in these long-tail settings (Fig.~\ref{fig:motivation}(c)-(d)). 

% To our knowledge, data-efficient adaptation of pretrained driving VLAs to trucks in long-tail highway scenarios has not been systematically studied. This gap raises the central question of our work: \emph{can limited demonstrations adapt a driving VLA to better match expert truck trajectories in these settings?} Alpamayo's~\cite{nvidia2025alpamayo,nvidia2026alpamayo15} modular architecture makes this directly testable. Its large-scale pretraining motivates reusing the vision-language backbone, while the separate flow-matching Action Expert provides a natural point for targeted refinement. We therefore keep the backbone frozen throughout our study. This design does not assume the pretrained representation is inherently truck-optimal; rather, it isolates how much of the observed prediction gap can be addressed strictly through action-side adaptation.

To our knowledge, data-efficient adaptation of pretrained driving VLAs to trucks in long-tail highway scenarios has not been systematically studied. 
% To address this gap, we leverage Alpamayo's modular architecture consisting of two major pieces: an 8.2B parameter Cosmos-Reason2 VLM backbone and a 2.3B parameter Action Expert. 
To address this gap, we leverage Alpamayo's modular architecture, which consists of an 8.2B-parameter Cosmos-Reason2 vision-language backbone and a 2.3B-parameter flow-matching Action Expert.
Its large-scale pretraining motivates reusing the vision-language backbone, while the separate flow-matching Action Expert provides a natural point for targeted refinement~\cite{nvidia2025alpamayo,nvidia2026alpamayo15}. 
% We keep the backbone frozen throughout to isolate how much of the observed prediction gap can be addressed strictly through action-side adaptation. 
We keep the backbone frozen throughout to isolate how much of the
observed prediction gap can be addressed through action-side
adaptation alone.
This setup motivates our central question: 
\emph{can limited targeted demonstrations adapt a driving VLA to better match expert truck trajectories in these settings without retraining its vision-language backbone?}

% \emph{can carefully chosen limited targeted demonstrations adapt a driving VLA to better match expert truck trajectories in these settings without retraining its vision-language backbone?}

To answer this question, we propose an \emph{adapt-then-steer} strategy. In Stage~I, the \emph{adapt} stage, we fine-tune only the action-generation stack on real-world truck scenarios from the target families, yielding \emph{Targeted Truck-SFT}. Stage~II, the \emph{steer} stage, asks whether the remaining trajectory error can be further reduced while keeping the adapted VLA fixed. During generation, the Action Expert constructs an action sequence through iterative updates governed by a predicted action-space flow velocity, which describes how the action sequence evolves and is distinct from the truck's physical velocity~\cite{nvidia2025alpamayo,lipman2023flow}.
This iterative generation process offers a way to refine predictions without further updating the adapted Action Expert: corrections to an intermediate action sequence can influence its subsequent predictions. To this end,
% To refine this process, 
we introduce \emph{Flow Velocity Steering} (FVS), a compact, flow-time-conditioned residual module that adds a learned correction to this velocity at each generation step. The corrected intermediate sequence is passed back to the frozen Action Expert, allowing earlier corrections to influence subsequent action generation.
We train FVS through steered rollouts using the same targeted demonstrations as Stage~I, exposing it to intermediate action sequences induced by its own earlier corrections.

We evaluate this strategy on a scenario-disjoint test set of construction and accident-related scenarios.
For single-candidate prediction, Targeted Truck-SFT reduces
average displacement error (ADE) and final displacement error (FDE)
over the full \(6.4\,\mathrm{s}\) horizon by approximately \(56\%\)
relative to the off-the-shelf VLA.
Using the same targeted demonstrations, FVS further reduces
Targeted Truck-SFT's full-horizon ADE and FDE by \(13.9\%\)
and \(16.5\%\), respectively.
% Finally, we examine the data efficiency of Stage~I by comparing
% targeted and general truck-driving supervision.
% Crucially, this adaptation is highly data-efficient, an important practical benefit given that routine truck-driving logs accumulate at scale whereas curated long-tail demonstrations remain scarce. At matched scenario counts, targeted supervision yields
% \(19\)-\(26\%\) lower full-horizon ADE than general truck-driving
% supervision. Targeted Truck-SFT also remains competitive with
% General Truck-SFT trained on approximately \(65\times\) as many
% scenarios.
% Crucially, our Stage~I comparisons show that targeted adaptation
% is highly data-efficient, 
% an important benefit given
% curated long-tail demonstrations remain scarce.
Crucially, our Stage~I comparisons show that targeted adaptation is highly data-efficient, an important benefit given the scarcity of curated long-tail demonstrations.
At matched data budgets, targeted supervision yields
\(19\)-\(26\%\) lower full-horizon ADE than general truck-driving
supervision. 
% Targeted Truck-SFT also remains competitive with
% General Truck-SFT trained on approximately \(65\times\) as many scenarios.
% Targeted Truck-SFT also remains competitive with General Truck-SFT, trained on approximately \(65\times\) as many truck-driving scenarios not specifically selected from the two target long-tail families.
Targeted Truck-SFT also remains competitive with a version of the same base VLA fine-tuned on approximately \(65\times\) as many general truck-driving scenarios, which were not specifically selected from the two target long-tail families.
Together, these results support our \emph{adapt-then-steer} strategy
for vehicle-domain transfer: preserve the pretrained vision-language backbone,
target action generation with scenario-aligned demonstrations,
and refine trajectories by steering the generative rollout procedure.
% Our evaluations measure open-loop agreement with logged expert
% trajectories; they do not establish closed-loop feasibility or
% deployment safety.

Our contributions are the following:
\begin{itemize}
    % \item We demonstrate Class~8 truck adaptation by fine-tuning only
    % the Action Expert, substantially reducing open-loop trajectory error
    % relative to off-the-shelf Alpamayo~1.5 while preserving the
    % vision-language backbone.
    % \item We demonstrate that fine-tuning only the action-generation stack closes a
    % substantial portion of off-the-shelf Alpamayo~1.5's open-loop error

    % \item We demonstrate that, with the vision-language backbone frozen,
    % fine-tuning only the action-generation stack closes a substantial
    % portion of the open-loop error gap between off-the-shelf
    % Alpamayo~1.5 predictions and logged expert truck trajectories.
    
\item We demonstrate that, with the vision-language backbone frozen, fine-tuning only the action-generation stack on targeted truck demonstrations reduces full-horizon trajectory error by approximately $56\%$ relative to the off-the-shelf VLA.

% \item We introduce FVS, a compact, flow-time-conditioned residual module that adds learned corrections to the action-space flow velocity at each generation step, complementing the fine-tuning process while keeping the adapted VLA fixed.

\item We introduce FVS, a compact, flow-time-conditioned residual module that adds learned corrections to the action-space flow velocity at each generation step. FVS complements the fine-tuning process by further reducing full-horizon ADE and FDE by $13.9\%$ and $16.5\%$, respectively, while keeping the adapted VLA fixed.
    % \item We show that targeted supervision yields lower full-horizon trajectory error at matched scenario counts than general truck-driving supervision, while Targeted Truck-SFT remains competitive with a General Truck-SFT model fine-tuned on approximately 65x as many scenarios.

    % \item We show that targeted supervision yields lower full-horizon trajectory error at matched scenario counts than general truck-driving supervision, and remains competitive with a model fine-tuned on approximately \(65\times\) as many general scenarios.
    \item  We show that targeted supervision yields lower full-horizon trajectory error at matched scenario counts than general truck-driving supervision, while the targeted model remains competitive with one fine-tuned on approximately \(65\times\) as many general scenarios.

\end{itemize}
% \input{sections/Intro_2}

%===============================================================================
\section{Related Work}
\label{sec:related_work}

\paragraph{Autonomous-driving VLAs and long-tail adaptation.}
Recent autonomous-driving VLAs extend multimodal scene understanding to trajectory generation. DriveVLM uses vision-language reasoning for scene analysis and
hierarchical planning, while EMMA predicts planner trajectories and other
driving outputs within a unified multimodal model
\cite{tian2025drivevlm,hwang2024emma}. Alpamayo-R1 combines
Chain-of-Causation reasoning with a generative trajectory predictor for
long-tail driving situations \cite{nvidia2025alpamayo}. Recent work explores
several interfaces for adapting driving policies to challenging scenarios.
SteerVLA learns a control-aligned language interface through which a
high-level vision-language model guides a low-level driving policy
\cite{gao2026steervla}; CLAP optimizes soft prompts for a frozen policy while
preserving nominal behavior \cite{zhu2026clap}; and TakeVLA uses takeover
data, language supervision, and reinforcement post-training
\cite{gao2026takevla}. More generally, adapters and low-rank updates provide
parameter-efficient interfaces for specializing large pretrained backbones
\cite{houlsby2019adapters,hu2022lora}. 
% Our work studies a complementary
% design choice: we preserve the vision-language backbone, specialize the
% action-generation stack using real truck trajectories, and then refine the resulting truck-adapted policy while keeping the fine-tuned VLA fixed.
Our work focuses on action-side adaptation for vehicle-domain transfer:
we specialize the action-generation stack using real truck trajectories while preserving the pretrained vision-language backbone.

\paragraph{Autonomous trucking and vehicle-domain transfer.}
Truck autonomy introduces sensing, embodiment, and operating-domain
differences that are not fully represented by passenger-vehicle benchmarks.
MAN TruckScenes documents truck-specific challenges including elevated sensor
perspectives, trailer occlusions, and terminal environments
\cite{fent2024truckscenes}, while TruckDrive emphasizes the long sensing and
anticipation ranges required for highway-speed trucking
\cite{ghilotti2026truckdrive}. MVAdapt explicitly formulates a vehicle-domain gap and conditions features from a frozen scene encoder on vehicle properties to transfer an end-to-end policy across vehicle embodiments in simulation
\cite{oh2026mvadapt}. 
% Our work studies adaptation of an open-weight autonomous-driving VLA
% using real Class~8 truck trajectories from construction and
% accident-related scenarios.
Our work focuses on this domain gap by adapting an open-weight autonomous-driving VLA to Class~8 trucks using real-world trajectories from construction and accident-related scenarios.
% Within this setting, we compare targeted long-tail supervision with
% general truck-driving supervision to assess the value of scenario
% alignment. 
% These studies motivate evaluating transfer across vehicle platforms rather than assuming that a pretrained policy transfers unchanged.
% We study adaptation from real Class~8 construction and accident-related
% demonstrations and compare targeted long-tail supervision with general
% truck-driving supervision.

\paragraph{Residual adaptation of generative action policies.}
Residual policy learning augments a nominal controller or policy with a
learned correction and has been used to improve otherwise imperfect robot
policies \cite{silver2019residualpolicy,johannink2019residual}.
% Diffusion Policy generates action sequences through iterative denoising
% \cite{chi2023diffusionpolicy}, while flow matching learns vector fields
% whose integration generates samples \cite{lipman2023flow}. 
% Recent work adapts frozen
% generative policies through related action- and flow-space interventions.
Recent work adapts pretrained generative policies by modifying their generated actions or the action-generation process while keeping the original policy weights fixed.
% Recent work adapts pretrained policies by modifying their generated actions or action-generation process while keeping the original policy weights fixed.
% FlowCorrect learns gated low-rank corrections to intermediate flow
% velocities from sparse human interventions, using remaining-flow-time
% targets along edited rollouts \cite{welte2026flowcorrect}; 
% FlowCorrect uses gated low-rank adapters to correct intermediate flow
% velocities from sparse human interventions \cite{welte2026flowcorrect}. It is
% trained on edited flow rollouts using velocity targets defined to reach the
% human-corrected action over the remaining flow time.
FlowCorrect uses gated low-rank adapters to correct intermediate flow velocities from sparse human interventions, training on remaining-flow-time velocity targets 
\cite{welte2026flowcorrect}.
RFS combines residual action adaptation with
reinforcement learning over the initial latent distribution
\cite{su2026rfs}; GLOVES refines completed action chunks using a separately
learned expert flow \cite{sun2026gloves}; 
% and RL$^2$-VLA composes a
% lightweight auxiliary flow policy with a frozen VLA
% \cite{tan2026rl2vla}. 
and RL$^2$-VLA learns an offline-RL policy conditioned on Action
Expert latents and composes its flow velocity with that of a frozen
VLA \cite{tan2026rl2vla}.
% Like FlowCorrect, FVS corrects intermediate action-space flow
% velocities, but uses a separate residual network rather than internal
% low-rank adapters.
% % We train FVS offline from complete logged expert truck trajectories
% % using action-space correction targets and decoded-trajectory
% % supervision, while keeping the truck-adapted VLA fixed.
% We train FVS offline from logged expert truck demonstrations using
% intermediate action-space bridge supervision and rollout-level imitation
% supervision, exposing it to generative action states induced by its own
% earlier corrections while keeping the truck-adapted policy fixed.
Building on the broader residual learning paradigm, FVS corrects intermediate action-space flow velocities using a separate, lightweight residual network.
We train FVS offline from logged expert truck demonstrations using intermediate action-space bridge supervision and rollout-level imitation supervision, exposing it to generative action states induced by its own earlier corrections while keeping the fine-tuned policy fixed.
%===============================================================================
\section{Method}
\label{sec:method}

\begin{figure*}[t]
    \centering
    \includegraphics[width=0.98\textwidth]{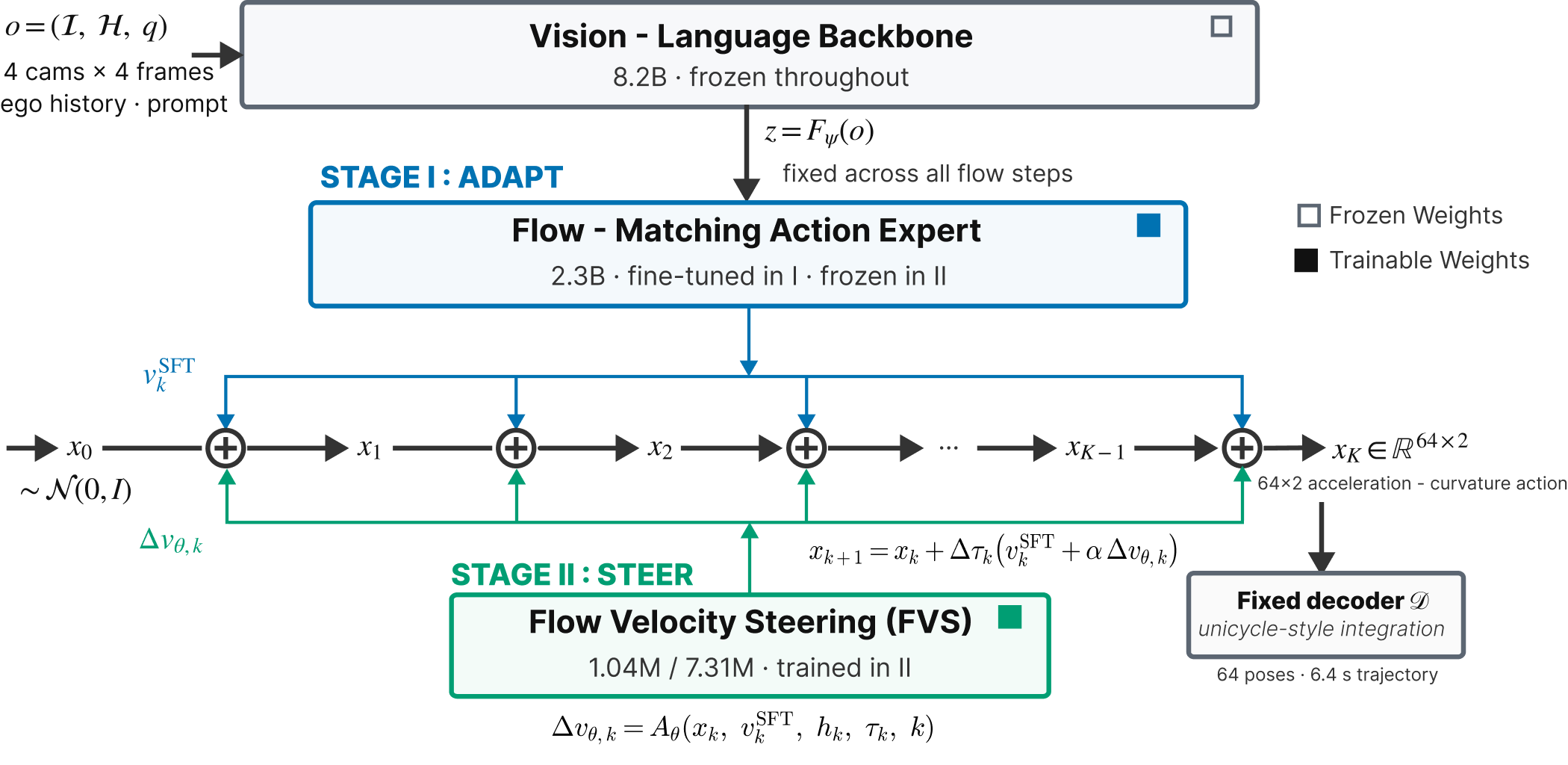}
   \caption{
Overview of the proposed adapt-then-steer pipeline.
Stage~I fine-tunes the action-generation stack with the vision-language backbone
frozen; Stage~II freezes the adapted policy and trains FVS to correct
its action-space flow velocity throughout the Euler rollout before fixed
trajectory decoding.
}
\vspace{-1.0em}

    \label{fig:method_overview}
\end{figure*}

\subsection{Problem Formulation and Overview}
\label{sec:problem}

We adapt a pretrained driving VLA model to
generate candidate trajectories for a Class~8 truck platform. Given an
observation $o=(\mathcal{I},\mathcal{H},q)$, where $\mathcal{I}$
denotes the synchronized multi-camera observation history,
$\mathcal{H}$ the recent ego-motion history, and $q$ the trajectory
generation prompt, the target
$Y^{\star}=\{(p_i^{\star},R_i^{\star})\}_{i=1}^{N}$ is the logged
future pose trajectory of the truck's vehicle control point (VCP),
where $p_i^{\star}$ and $R_i^{\star}$ denote its future position and
orientation. The VCP serves as the common reference point for both
logged and predicted truck motion, and each target trajectory is
expressed in the coordinate frame of the current VCP pose.

% Following Alpamayo~1.5
% \cite{nvidia2026alpamayo15,nvidia2025alpamayo}, the policy operates
% in its native normalized acceleration-curvature action space
% rather than directly predicting Cartesian poses. We retain Alpamayo's fixed
% trajectory-to-action mapping $\mathcal{T}$ and action-to-trajectory
% decoder $\mathcal{D}$ unchanged. The decoder maps the generated
% longitudinal-acceleration-curvature sequence to a VCP pose trajectory
% through Alpamayo's unicycle-style integration from the current vehicle
% state:
% Following Alpamayo~1.5
% \cite{nvidia2026alpamayo15,nvidia2025alpamayo}, we retain its
% native normalized acceleration-curvature action representation,
% fixed trajectory-to-action mapping $\mathcal{T}$, and fixed
% decoder $\mathcal{D}$. The decoder integrates actions from the
% current vehicle state using Alpamayo's unicycle-style model:
Following Alpamayo~1.5
\cite{nvidia2026alpamayo15,nvidia2025alpamayo}, we retain its
native normalized acceleration-curvature action representation
and associated fixed conversion routines. The mapping
$\mathcal{T}$ constructs action targets from logged trajectories,
while $\mathcal{D}$ denormalizes and integrates generated
actions using the native unicycle model:
\begin{equation}
\begin{aligned}
    x^{\star}
    &=
    \mathcal{T}(Y^{\star},\mathcal{H})
    \in \mathbb{R}^{N\times 2},
    \\
    \widehat{Y}
    &=
    \mathcal{D}(x_K,\mathcal{H}),
\end{aligned}
\label{eq:action_representation}
\end{equation}
% where $x^{\star}$ is the expert truck action sequence and $x_K$ is
% the generated action after $K$ flow-integration steps. The trajectory
% horizon, waypoint spacing, and action-space configuration are provided
% % in Sec.~\ref{sec:experimental_setup}.
% where $x^{\star}$ is the expert truck action sequence and $x_K$ is
% the generated action after $K$ flow-integration steps. The decoder
% remains unchanged throughout both adaptation stages and is
% differentiable during FVS training. The trajectory horizon, waypoint
% spacing, number of flow-integration steps, and action-space
% configuration are provided in Sec.~\ref{sec:experimental_setup}.
% Here and throughout Sec.~\ref{sec:method}, $K$ denotes the number of
% internal flow-integration steps; the number of independently sampled
% candidate trajectories is denoted separately by $K_{\mathrm{cand}}$
% in Sec.~\ref{sec:experimental_setup}.
where $x^{\star}$ is the expert truck action sequence and $x_K$ is
the generated action after $K$ flow-integration steps.
The fixed decoder is differentiable with respect to the
generated actions during FVS training.
Trajectory-generation settings are given in
Sec.~\ref{sec:experimental_setup}.

% Our adaptation proceeds in two stages, as summarized in
% Fig.~\ref{fig:method_overview}. In Stage~I, we freeze the
% vision-language backbone and perform supervised fine-tuning of the
% action-generation stack on logged truck demonstrations. In Stage~II,
% we freeze the complete truck-SFT policy and learn a compact,
% flow-time-conditioned residual module that steers its action-space
% velocity field throughout the Euler rollout. We refer to this second
% stage as \emph{Flow Velocity Steering} (FVS). Thus, SFT specializes
% the policy's conditional action field to the truck data, while FVS
% selectively refines the remaining trajectory errors without further
% updating the VLM or action expert.

% Our adaptation proceeds in two stages
% (Fig.~\ref{fig:method_overview}).
% Stage~I fine-tunes the action-generation stack on logged truck
% demonstrations while keeping the vision-language backbone frozen.
% Stage~II freezes the resulting policy and learns
% \emph{Flow Velocity Steering} (FVS), a flow-time-conditioned residual
% that corrects the Action Expert's action-space flow velocity
% throughout the Euler rollout.
Our adaptation proceeds in two stages
(Fig.~\ref{fig:method_overview}). Stage~I fine-tunes the
action-generation stack with the vision-language backbone
frozen. Stage~II freezes the adapted policy and learns FVS
to correct its action-space flow velocity throughout generation.

\subsection{Truck-Specific Action-Expert Fine-Tuning}
\label{sec:sft}
Let $F_{\psi}$ denote the pretrained vision-language backbone and
$z=F_{\psi}(o)$ its multimodal context. 
% During SFT, the backbone
% parameters $\psi$ remain frozen, while the action-generation stack, parameterized by $\phi$ and comprising the Action Expert and its
% input/output projections, is fine-tuned on the truck demonstrations.
During SFT, the backbone parameters $\psi$ remain frozen, while the flow-matching action stack, parameterized by $\phi$ and comprising the
Action Expert and its action input/output projections, is fine-tuned
on the truck demonstrations.

We retain Alpamayo~1.5's native conditional flow-matching
objective~\cite{nvidia2026alpamayo15,nvidia2025alpamayo,lipman2023flow}. For an expert truck action sequence $x^{\star}$, we sample
Gaussian noise $\epsilon\sim\mathcal{N}(0,I)$ and generative
flow time $\tau\in[0,1]$ using the native SFT sampling
distribution. We construct
$x_{\tau}=(1-\tau)\epsilon+\tau x^{\star}$ and optimize $\phi$ using
% Let $F_{\psi}$ denote the pretrained vision-language backbone and
% $z=F_{\psi}(o)$ its multimodal context. During SFT, the backbone
% parameters $\psi$ remain frozen, while the action-generation stack,
% parameterized by $\phi$, is fine-tuned on the truck demonstrations.

% We use Alpamayo~1.5's native conditional flow-matching
% objective~\cite{nvidia2026alpamayo15,nvidia2025alpamayo,lipman2023flow}. For an expert
% truck action $x^{\star}$, we sample Gaussian noise
% $\epsilon\sim\mathcal{N}(0,I)$ and generative flow time
% $\tau\in[0,1]$, and construct
% $x_{\tau}=(1-\tau)\epsilon+\tau x^{\star}$. The action expert is
% optimized using
\begin{equation}
    \mathcal{L}_{\mathrm{SFT}}
    =
    \mathbb{E}
    \left[
        \left\|
            v_{\phi}(x_{\tau},\tau,z)
            -
            (x^{\star}-\epsilon)
        \right\|_{F}^{2}
    \right],
    \label{eq:sft_loss}
\end{equation}
where $v_{\phi}$ is the predicted action-space flow velocity.
% At inference, sampling starts from
% $x_0\sim\mathcal{N}(0,I)$ and follows the native Euler solver,
At inference, sampling starts from
$x_0\sim\mathcal{N}(0,I)$ and follows the native Euler solver
with step sizes $\Delta\tau_k=\tau_{k+1}-\tau_k$:
\begin{equation}
    x_{k+1}
    =
    x_k+\Delta\tau_k\,
    v_{\phi}(x_k,\tau_k,z),
    \qquad k=0,\ldots,K-1.
    \label{eq:native_flow}
\end{equation}
% The final action is decoded as
% $\widehat{Y}=\mathcal{D}(x_K,\mathcal{H})$. Here, $\tau$ denotes
% generative flow time; the flow velocity is a direction in normalized
% action space, not the physical velocity of the truck.
% The final action is decoded as
% $\widehat{Y}=\mathcal{D}(x_K,\mathcal{H})$. Here, $\tau$ denotes
% generative flow time: the scene context remains fixed while the
% action state $x_k$ evolves through the generative rollout. 
% \satya{The flow velocity describes the rate of change of the normalized
% action state with respect to $\tau$, not the physical velocity
% of the truck.}
% The final action is decoded as
% $\widehat{Y}=\mathcal{D}(x_K,\mathcal{H})$. Here, $\tau$ denotes
% generative flow time: the scene context remains fixed while the
% action state $x_k$ evolves through the generative rollout.
% The flow velocity describes the rate of change of the normalized
% action state with respect to $\tau$, not the physical velocity
% of the truck.
During sampling, the context $z$ remains fixed while $x_k$
evolves in generative flow time $\tau$. Flow velocity describes
the rate of change of normalized actions with respect to
$\tau$, not the truck's physical velocity.

\subsection{Flow Velocity Steering}
\label{sec:fvs}

% FVS acts during the adapted policy's iterative
% action-generation process rather than correcting only the
% completed action sequence.
% While the native SFT objective in Eq.~(\ref{eq:sft_loss}) fits action-space flow velocities,
% it does not directly supervise the trajectory decoded from a
% complete generated action sequence.
% To address this, we freeze the complete adapted policy after SFT. At Euler step $k$, the frozen Action Expert is evaluated at
% the current steered action state $x_k$, returning the
% action-space flow velocity
% $v_k^{\mathrm{SFT}} = v_\phi(x_k,\tau_k,z)$
% and its hidden action representation $h_k$.
% FVS predicts a residual correction to the frozen policy's
% flow velocity:
% \begin{equation}
%     \Delta v_{\theta,k}
%     =
%     A_\theta
%     \left(
%         x_k,
%         v_k^{\mathrm{SFT}},
%         h_k,
%         \tau_k,
%         k
%     \right),
%     \label{eq:fvs_residual}
% \end{equation}
% The steered sampler then uses
% \begin{equation}
%     x_{k+1}
%     =
%     x_k
%     +
%     \Delta\tau_k
%     \left(
%         v_k^{\mathrm{SFT}}
%         +
%         \alpha\,\Delta v_{\theta,k}
%     \right),
%     \label{eq:steered_flow}
% \end{equation}
% where $\alpha$ controls the steering strength; $\alpha=0$
% recovers the frozen SFT sampler.
% Because the resulting intermediate action sequence is passed
% back to the frozen Action Expert at the next Euler step,
% FVS influences subsequent generation rather than modifying
% only the completed action sequence.

FVS acts during the adapted policy's iterative
action-generation process rather than correcting only the
completed action sequence.
% While the native SFT objective in Eq.~(\ref{eq:sft_loss}) fits action-space flow velocities,
% it does not directly supervise the trajectory decoded from a
% complete generated action sequence.
While the native SFT objective in  Eq.~(\ref{eq:sft_loss}) fits action-space
flow velocities, it does not explicitly supervise the
trajectory obtained after a complete generative rollout
and decoding.
This motivates a complementary refinement stage that
directly supervises the generated trajectory while updating
only a compact residual module.
To address this, we freeze the adapted policy after SFT and 
introduce FVS as a trainable residual module conditioned on flow time.
This allows us to supervise both intermediate corrections and the 
final decoded trajectory through complete generative rollouts.

Specifically, at each Euler step $k$, the frozen Action Expert is evaluated at
the current steered action state $x_k$, returning the nominal
action-space flow velocity
$v_k^{\mathrm{SFT}} = v_\phi(x_k,\tau_k,z)$
and its hidden action representation $h_k$.
FVS then predicts a residual correction to this velocity:
\begin{equation}
    \Delta v_{\theta,k}
    =
    A_\theta
    \left(
        x_k,
        v_k^{\mathrm{SFT}},
        h_k,
        \tau_k,
        k
    \right).
    \label{eq:fvs_residual}
\end{equation}
The steered sampler applies this correction to update the action state:
\begin{equation}
    x_{k+1}
    =
    x_k
    +
    \Delta\tau_k
    \left(
        v_k^{\mathrm{SFT}}
        +
        \alpha\,\Delta v_{\theta,k}
    \right),
    \label{eq:steered_flow}
\end{equation}
where $\alpha$ controls the steering strength; $\alpha=0$
recovers the frozen SFT sampler.
Because the resulting intermediate action sequence is passed
back to the frozen Action Expert at the next Euler step,
FVS dynamically steers the entire rollout.

\subsection{Trajectory-Aware Steering Objective}
\label{sec:fvs_objective}

We train FVS from the same logged expert demonstrations used for SFT, using action-space bridge supervision at intermediate flow steps and imitation supervision on complete steered rollouts.
For each training example, the frozen SFT and steered rollouts
share the same initial Gaussian noise.
The matched frozen rollout provides a training-only difficulty
weight $\eta\in[0,1]$, derived from its trajectory error using
training-set statistics; larger errors yield higher weights.

We construct a bridge target to guide the current steered
action sequence toward the expert action.
Let $R_k=\sum_{j=k}^{K-1}\Delta\tau_j$ denote the remaining
flow time. Over this interval, the constant action-space flow
velocity $(x^{\star}-x_k)/R_k$ would move the current action
state $x_k$ to the expert action $x^{\star}$.
Since FVS acts as a residual on the frozen Action Expert's
action-space flow velocity, we define the bridge target
\begin{equation}
    \Delta v_k^{\star}
    =
    \eta\,\beta
    \left(
        \frac{x^{\star}-x_k}
             {\max(R_k,R_{\min})}
        -
        v_k^{\mathrm{SFT}}
    \right),
    \label{eq:bridge_target}
\end{equation}
where $R_{\min}$ stabilizes the denominator and $\beta$
controls the target scale.
The bridge loss $\mathcal{L}_{\mathrm{bridge}}$ penalizes
deviation of the predicted correction $\Delta v_{\theta,k}$
from the bridge target $\Delta v_k^{\star}$.

After the complete steered rollout, we decode
$\widehat{Y}_{\theta}
=\mathcal{D}(x_K^{\theta},\mathcal{H})$
and compute a per-example imitation loss $E_{\theta}$
supervising the generated action sequence and decoded trajectory
against the logged expert demonstration.
Let $E_{\mathrm{SFT}}$ denote the same loss evaluated on the
matched-noise frozen SFT prediction.
We use $\mathcal{L}_{\mathrm{traj}}=E_{\theta}$ and penalize
regression relative to the frozen policy using
$\mathcal{L}_{\mathrm{nr}}
=[E_{\theta}-E_{\mathrm{SFT}}]_{+}$,
where $[a]_{+}=\max(a,0)$.

The complete objective is
\begin{equation}
    \mathcal{L}_{\mathrm{FVS}}
    =
    \lambda_b\mathcal{L}_{\mathrm{bridge}}
    +
    \lambda_t\mathcal{L}_{\mathrm{traj}}
    +
    \lambda_{\mathrm{nr}}\mathcal{L}_{\mathrm{nr}}
    +
    \mathcal{L}_{\mathrm{reg}},
    \label{eq:fvs_loss}
\end{equation}
where $\mathcal{L}_{\mathrm{reg}}$ collects correction-magnitude
and architecture-specific regularization terms.

\paragraph{Rollout-aware optimization}
We optimize FVS through complete steered rollouts under each
fixed logged observation.
Because each correction modifies the action state $x_k$,
earlier FVS outputs affect the states encountered at later
flow steps.
FVS is therefore trained on action states induced by its own
corrections rather than only on frozen-policy states, reducing
the resulting state-distribution mismatch
\cite{ross2011dagger}.
The adapted VLA remains fixed throughout, and training requires
neither environment interaction nor additional expert queries.

\section{Experimental Results}
\label{sec:experimental_results}

\begin{table*}[t]
    \centering
   
    \setlength{\tabcolsep}{3.2pt}
    \resizebox{\textwidth}{!}{
    \begin{tabular}{lcccccccc}
        \toprule
        &
        \multicolumn{4}{c}{\(K_{\mathrm{cand}}=1\): single candidate}
        &
        \multicolumn{4}{c}{\(K_{\mathrm{cand}}=16\): oracle best-of-16}
        \\
        \cmidrule(lr){2-5}
        \cmidrule(lr){6-9}
        Method
        & ADE@1s
        & ADE@3s
        & ADE@6.4s
        & FDE@6.4s
        & minADE@1s
        & minADE@3s
        & minADE@6.4s
        & minFDE@6.4s
        \\
        \midrule

        CTRV
        & \(0.1643\)
        & \(0.7003\)
        & \(2.0430\)
        & \(5.8412\)
        & \textemdash
        & \textemdash
        & \textemdash
        & \textemdash
        \\

        Base
        & \(0.1530 \pm 0.0018\)
        & \(0.6836 \pm 0.0440\)
        & \(2.6358 \pm 0.1493\)
        & \(7.1830 \pm 0.2868\)
        & \(0.0621 \pm 0.0001\)
        & \(0.2556 \pm 0.0082\)
        & \(0.9402 \pm 0.0038\)
        & \(2.3849 \pm 0.0118\)
        \\

        \midrule
        Targeted Truck-SFT
        & \(0.0600 \pm 0.0008\)
        & \(0.3120 \pm 0.0081\)
        & \(1.1650 \pm 0.0184\)
        & \(3.1590 \pm 0.0213\)
        & \(0.0413 \pm 0.0004\)
        & \(0.1855 \pm 0.0029\)
        & \(0.6968 \pm 0.0115\)
        & \(1.8118 \pm 0.0398\)
        \\

       General Truck-SFT
        & \(0.0584 \pm 0.0017\)
        & \(0.3146 \pm 0.0194\)
        & \(1.2868 \pm 0.0049\)
        & \(3.7404 \pm 0.0824\)
        & \(0.0361 \pm 0.0001\)
        & \(0.1659 \pm 0.0051\)
        & \(0.6634 \pm 0.0044\)
        & \(1.7652 \pm 0.0240\)
        \\

        General$\rightarrow$Targeted SFT
        & \(0.0580 \pm 0.0017\)
        & \(0.3121 \pm 0.0168\)
        & \(1.2539 \pm 0.0422\)
        & \(3.6012 \pm 0.0703\)
        & \(0.0371 \pm 0.0008\)
        & \(0.1716 \pm 0.0045\)
        & \(0.6697 \pm 0.0009\)
        & \(1.7587 \pm 0.0313\)
        \\

        \midrule
        Final-Action Residual (MLP)
        & \(0.0595 \pm 0.0006\)
        & \(0.3060 \pm 0.0043\)
        & \(1.1362 \pm 0.0092\)
        & \(3.1071 \pm 0.0186\)
        & \(0.0402 \pm 0.0003\)
        & \(0.1813 \pm 0.0012\)
        & \(0.6918 \pm 0.0073\)
        & \(1.8001 \pm 0.0232\)
        \\

        Final-Action Residual (Transformer)
        & \(0.0589 \pm 0.0006\)
        & \(0.3010 \pm 0.0032\)
        & \(1.1276 \pm 0.0081\)
        & \(3.0002 \pm 0.0175\)
        & \(0.0391 \pm 0.0003\)
        & \(0.1801 \pm 0.0016\)
        & \(0.6876 \pm 0.0051\)
        & \(1.7992 \pm 0.0210\)
        \\

        FVS (MLP)
        & \(0.0481 \pm 0.0003\)
        & \(0.2659 \pm 0.0004\)
        & \(1.0065 \pm 0.0035\)
        & \(2.6527 \pm 0.0171\)
        & \(0.0363 \pm 0.0000\)
        & \(0.1658 \pm 0.0001\)
        & \(0.6613 \pm 0.0027\)
        & \(1.7419 \pm 0.0115\)
        \\

        FVS (Transformer)
        & \(\mathbf{0.0468 \pm 0.0003}\)
        & \(\mathbf{0.2656 \pm 0.0009}\)
        & \(\mathbf{1.0035 \pm 0.0035}\)
        & \(\mathbf{2.6383 \pm 0.0156}\)
        & \(\mathbf{0.0354 \pm 0.0000}\)
        & \(\mathbf{0.1642 \pm 0.0006}\)
        & \(\mathbf{0.6573 \pm 0.0012}\)
        & \(\mathbf{1.7003 \pm 0.0102}\)
        \\

        \bottomrule
    \end{tabular}
    }
\caption{
Trajectory accuracy on the 26-scenario held-out truck long-tail test set.
\(K_{\mathrm{cand}}=1\) evaluates one candidate, whereas
\(K_{\mathrm{cand}}=16\) reports oracle best-of-16
\(\operatorname{minADE}/\operatorname{minFDE}\).
Learned-model values are mean \(\pm\) sample standard deviation across
inference-noise seeds.
CTRV is deterministic and reported only for \(K_{\mathrm{cand}}=1\).
Errors are in metres; lower is better.
Bold indicates the lowest mean in each column.
Final-action residual and FVS models use the same frozen
Targeted Truck-SFT checkpoint.
}
\vspace{-.5em}

\label{tab:main_results}
\end{table*}

\subsection{Dataset and Evaluation Protocol}
\label{sec:experimental_setup}

\paragraph{Datasets and splits}

We study two target long-tail scenario families: construction zones and
accident-related road disruptions.
Each targeted scenario records an expert truck driver's encounter
with one of these situations, including the maneuver through it.
Multiple prediction windows are sampled within each scenario.
The targeted training set contains 229 scenarios, comprising
138 accident-related and 91 construction scenarios.
The general training set contains 14,891 truck-driving scenarios not
specifically selected for these families.
Both corpora contain real truck-driving data.
% All primary methods are evaluated on the same scenario-disjoint held-out
% test set of 26 construction and accident-related scenarios.
% All primary methods are evaluated on the same held-out test set of 26 construction and accident-related scenarios, which is
% scenario-disjoint from both the targeted and general training corpora.
All primary methods are evaluated on the same
scenario-disjoint held-out set of 26 real-world truck
encounters spanning construction zones and accident-related
road disruptions. 
% Multiple prediction windows are sampled
% within each encounter, and each scenario contributes equally
% to the aggregate metrics.
% Checkpoints and hyperparameters were selected without reference to the
% held-out test scenarios.
The test scenarios are disjoint from both the targeted and
general training corpora, and checkpoints and hyperparameters
were selected without reference to them.

\paragraph{Compared methods}
\emph{Base} denotes the off-the-shelf Alpamayo~1.5 checkpoint.
\emph{Targeted Truck-SFT} and \emph{General Truck-SFT} fine-tune the
action-generation stack on the targeted and general training sets,
respectively, while preserving the vision-language backbone.
\emph{General$\rightarrow$Targeted SFT} initializes from General
Truck-SFT and is subsequently fine-tuned on the targeted scenarios.

% For post-SFT refinement, FVS and the \emph{final-action residual}
% baseline both start from the same frozen Targeted Truck-SFT checkpoint.
% The final-action residual is conditioned on the frozen Action Expert's
% final projected hidden representation and applies a scaled, bounded
% correction to the completed acceleration-curvature sequence before
% trajectory decoding, leaving the native SFT flow rollout unchanged.
% It is trained using action-space residual and decoded-trajectory losses
% together with residual-magnitude regularization.
For post-SFT refinement, FVS and the \emph{final-action residual}
baseline both start from the same frozen Targeted Truck-SFT checkpoint.
% The final-action residual applies a learned correction to the completed
% acceleration-curvature sequence before trajectory decoding, whereas
% FVS applies residual corrections during iterative action generation.
% FVS instead corrects the action-space flow velocity throughout
% generation, applying its correction at every Euler step rather than to the completed sequence.
The final-action residual applies a learned correction to the completed acceleration-curvature sequence before trajectory decoding, whereas FVS corrects the action-space flow velocity at every Euler step during generation.
We instantiate both approaches with MLP and Transformer residual
networks.
Within each architecture, the two approaches use the same targeted
training set, optimizer settings, training-update budget, and
trainable-parameter count.
The MLP and Transformer residual modules contain \(1.044\)M and
\(7.308\)M trainable parameters, respectively.
Each FVS module therefore adds less than \(0.1\%\) to the parameter
count of the complete frozen VLA.
% We also test whether truck-specific instructions improve Base without
% weight updates. The prompts incorporate truck-driver feedback on turning radius,
% trailer off-tracking, braking, clearance, and smooth control.

% To assess whether adaptation improves on extrapolating recent truck
% motion, we include a deterministic constant-turn-rate-and-velocity
% (CTRV) baseline \cite{schubert2008motionmodels}.
% CTRV uses the same full causal ego-motion history supplied to Base.
% For each adjacent pose pair, we estimate signed longitudinal speed
% by projecting displacement onto the angular midpoint heading and
% yaw rate from the wrapped heading difference, using the actual
% time interval.
% The median speed and median yaw rate across the full history are
% held constant to analytically propagate a trajectory from the current
% pose, using the same prediction horizon and sampling interval as
% the learned models.
% We use the straight-line limit for yaw-rate magnitudes below
% \(10^{-6}\,\mathrm{rad/s}\).
As a deterministic motion-only reference, we include a
constant-turn-rate-and-velocity (CTRV) baseline
\cite{schubert2008motionmodels}.
CTRV uses the full causal ego-motion history also supplied to Base.
For each adjacent pose pair, we estimate signed longitudinal speed by
projecting the displacement onto the midpoint heading and estimate yaw
rate from the wrapped heading difference, in both cases using the
actual time interval.
We then hold the median speed and median yaw rate over the full history
constant and analytically propagate a trajectory from the current pose,
using the same prediction horizon and sampling interval as the learned
models.
\paragraph{Data-efficiency protocol}
For each evaluated training-set budget, scenarios are randomly sampled
from the targeted and general corpora.
We train three independently sampled subset runs for each supervision
source at each budget.
All subset runs, together with the full-data Targeted Truck-SFT and
General Truck-SFT runs, use the same training-update budget.
The selected checkpoint from each run is evaluated on the common
held-out test set.
Figure~\ref{fig:data_sweep} reports the mean and standard deviation
across the three subset runs.

\paragraph{Trajectory generation}
Each prediction uses four synchronized truck-mounted camera streams
with four frames per camera, preserving Alpamayo~1.5's default
four-camera, four-frame-per-camera input count. The truck rig provides
corresponding camera views, with mounting positions adapted to the
Class~8 vehicle embodiment.
Each learned candidate is generated as a sequence of 64 pairs of
normalized longitudinal acceleration and curvature using \(K=10\)
Euler flow-integration steps.
The fixed Alpamayo decoder maps this sequence to a
\(6.4\,\mathrm{s}\) VCP trajectory containing 64 future poses at
\(0.1\,\mathrm{s}\) intervals.
Candidate generation is invoked at \(1\,\mathrm{Hz}\) along each
scenario, so successive prediction horizons overlap.
All evaluations are open-loop: each prediction is conditioned on
the logged camera observations and ego-motion history at its timestamp,
and is scored against the logged expert trajectory over the same horizon.
% A validated closed-loop visual simulator matching both our truck
% sensor configuration and articulated vehicle dynamics was unavailable
% for this study.
% Each learned candidate is generated as a sequence of 64 pairs of
% normalized longitudinal acceleration and curvature using \(K=10\)
% Euler flow-integration steps.
% The fixed Alpamayo decoder maps this sequence to a
% \(6.4\,\mathrm{s}\) VCP trajectory containing 64 future poses at
% \(0.1\,\mathrm{s}\) intervals.
% Candidate generation is invoked at \(1\,\mathrm{Hz}\) along each
% scenario, so successive prediction horizons overlap.
% All reported evaluations therefore measure open-loop agreement with
% logged truck trajectories.
% A validated closed-loop visual simulation setup matching both our truck
% sensor configuration and articulated vehicle dynamics was unavailable
% for this study.

\paragraph{Metrics}
% We denote the number of independently sampled candidate trajectories by
% \(K_{\mathrm{cand}}\), distinct from the number of flow steps \(K\).
% For \(K_{\mathrm{cand}}=1\), average displacement error (ADE) at horizon
% \(H\) is the mean Euclidean VCP-position error over the predicted poses
% through \(H\); final displacement error (FDE) is the position error at
% that horizon.
% We report ADE at \(1\), \(3\), and \(6.4\,\mathrm{s}\), and FDE at
% \(6.4\,\mathrm{s}\).
% Supporting geometry metrics at \(6.4\,\mathrm{s}\) comprise lateral and
% longitudinal ADE, defined as mean absolute VCP-position error along the
% corresponding coordinate axes, and final heading error, defined as the
% absolute angular difference between predicted and logged final orientations.
% Within each evaluation run, metrics are averaged over all held-out
% prediction windows. Each scenario contributes the same number of windows and therefore
% has equal weight in the aggregate.

We denote the number of independently sampled candidate trajectories by
\(K_{\mathrm{cand}}\), distinct from the number of flow-integration
steps \(K\).
For \(K_{\mathrm{cand}}=1\), ADE at horizon \(H\) is the mean Euclidean
VCP-position error over the predicted poses through \(H\), whereas FDE
is the VCP-position error at that horizon.
We report ADE at \(1\), \(3\), and \(6.4\,\mathrm{s}\), and FDE at
\(6.4\,\mathrm{s}\).
At \(6.4\,\mathrm{s}\), we additionally report lateral and longitudinal
ADE, defined as mean absolute VCP-position errors along the
corresponding coordinate axes, and final heading error, defined as the
absolute angular difference between the predicted and logged final
orientations.

For \(K_{\mathrm{cand}}=16\), we report oracle best-of-16
\(\operatorname{minADE}\) and \(\operatorname{minFDE}\).
For each prediction window, each metric independently selects its
lowest-error candidate before the per-window minima are averaged.
The selected candidate may therefore differ across metrics.
These oracle quantities measure best-of-set agreement with the logged
trajectory and do not assume an online selector with access to ground
truth.

For both candidate settings, metrics are averaged over all held-out
prediction windows.
Each scenario contributes the same number of windows and therefore has
equal weight in the aggregate.
Because each window is evaluated against one logged expert trajectory,
alternative valid maneuvers may incur nonzero displacement error.
For each learned-model checkpoint and inference-noise seed, metrics are
first aggregated over the common set of evaluation windows.
Tables~\ref{tab:main_results} and~\ref{tab:geometry_results} report the
mean and sample standard deviation of these per-seed aggregates.
The reported deviations therefore characterize inference-sampling
variability for fixed checkpoints and evaluation scenarios.

\subsection{Targeted Fine-Tuning for Truck Adaptation}
\label{sec:results_targeted_sft}
We compare Base with Targeted Truck-SFT trained on
229 targeted scenarios (Table~\ref{tab:main_results}).
With the vision-language backbone frozen, action-side
fine-tuning reduces full-horizon single-candidate ADE
and FDE by approximately \(56\%\), with ADE decreasing
from \(2.636\) to \(1.165\,\mathrm{m}\).
% We first compare Base with Targeted Truck-SFT trained on
% 229 targeted scenarios (Table~\ref{tab:main_results}).
% With the vision-language backbone frozen, fine-tuning the
% action-generation stack reduces full-horizon single-candidate
% ADE and FDE by approximately \(56\%\), with ADE decreasing
% from \(2.636\) to \(1.165\,\mathrm{m}\).
The CTRV baseline provides a motion-only reference by
extrapolating recent truck motion without visual or language
input.
Although Base has lower single-candidate ADE than CTRV at
\(1\) and \(3\,\mathrm{s}\), its full-horizon ADE and FDE
exceed those of CTRV, motivating truck-specific adaptation
of the off-the-shelf policy.
Targeted Truck-SFT achieves lower error than both Base
and CTRV at every reported single-candidate horizon.

% The improvement extends across the held-out scenarios:
% Targeted Truck-SFT lowers scenario-mean
% ADE@\(6.4\,\mathrm{s}\) relative to Base in all 26 scenarios.
% The mean percentage reductions are approximately \(63\%\)
% for accident-related scenarios and \(52\%\) for construction
% scenarios.
The improvement is consistent across the held-out set:
Targeted Truck-SFT lowers scenario-mean
ADE@\(6.4\,\mathrm{s}\) relative to Base in all 26 scenarios,
with mean reductions of approximately \(63\%\) for
accident-related scenarios and \(52\%\) for construction
scenarios.
Targeted Truck-SFT also lowers all reported oracle
best-of-16 errors relative to Base, improving best-of-set
agreement with logged truck trajectories.
The gains span multiple components of trajectory geometry:
lateral and longitudinal ADE are roughly halved, while
final heading error decreases from \(3.293^{\circ}\) to
\(0.885^{\circ}\) (Table~\ref{tab:geometry_results}).

% In a separate preliminary diagnostic using one inference-noise seed, we evaluated truck-specific prompts informed by
% truck-driver feedback on turning, trailer off-tracking,
% braking, clearance, and smooth control.
% None achieved lower single-candidate ADE@\(6.4\,\mathrm{s}\)
% than the default prompt on the same diagnostic windows.

% A parameter-free alternative to fine-tuning is to adapt the prompt rather than the model.
% Holding Base fixed, we substituted for the default prompt \(q\) a set of truck-specific prompts elicited from expert truck-driver feedback, describing turning behavior, trailer off-tracking, braking, clearance, and control smoothness.
% In a separate diagnostic on held-out events using one inference-noise seed, none reduced single-candidate ADE@\(6.4\,\mathrm{s}\) relative to the default prompt. All main learned-model comparisons use the default prompt \(q\) 
% (Sec.~\ref{sec:problem}).

As a parameter-free alternative to fine-tuning, we kept Base fixed and evaluated truck-specific alternatives to the default prompt $q$ (Sec.~\ref{sec:problem}). These prompts were informed by expert truck-driver feedback on turning behavior, trailer off-tracking, braking, clearance, and control smoothness. In a separate diagnostic on held-out events using one inference-noise seed, none of the tested prompts reduced single-candidate ADE@$6.4\,\mathrm{s}$ relative to the default prompt. Accordingly, all main learned-model comparisons use the default prompt $q$.

Together, these results show that action-side adaptation
substantially improves agreement with logged truck trajectories
while keeping the vision-language backbone frozen.

\begin{table}[t]
    \centering
    \setlength{\tabcolsep}{3.2pt}
    \renewcommand{\arraystretch}{1.08}
    \resizebox{\columnwidth}{!}{
    \begin{tabular}{lccc}
        \toprule
        Method
       & Lateral ADE \(\downarrow\)
& Longitudinal ADE \(\downarrow\)
& Heading Error \(\downarrow\)
        \\
        \midrule

        CTRV
        & \(0.7958\)
        & \(1.6934\)
        & \(1.6162\)
        \\

        Base
        & \(1.0204 \pm 0.0638\)
        & \(2.1706 \pm 0.2306\)
        & \(3.2926 \pm 0.8549\)
        \\

        \midrule
        Targeted Truck-SFT
        & \(0.4749 \pm 0.0130\)
        & \(0.9323 \pm 0.0197\)
        & \(0.8851 \pm 0.0232\)
        \\

        General Truck-SFT
        & \(0.4832 \pm 0.0152\)
        & \(1.0929 \pm 0.0402\)
        & \(1.1377 \pm 0.0459\)
        \\

        General$\rightarrow$Targeted SFT
        & \(0.4932 \pm 0.0140\)
        & \(1.0385 \pm 0.0298\)
        & \(1.1192 \pm 0.0553\)
        \\

        \midrule
        Final-Action Residual (MLP)
        & \(0.4634 \pm 0.0070\)
        & \(0.9115 \pm 0.0115\)
        & \(0.8590 \pm 0.0172\)
        \\

        Final-Action Residual (Transformer)
        & \(0.4592 \pm 0.0068\)
        & \(0.9009 \pm 0.0117\)
        & \(0.8512 \pm 0.0171\)
        \\

        FVS (MLP)
        & \(0.4059 \pm 0.0053\)
        & \(0.7738 \pm 0.0127\)
        & \(0.7644 \pm 0.0120\)
        \\

        FVS (Transformer)
        & \(\mathbf{0.3870 \pm 0.0051}\)
        & \(\mathbf{0.7650 \pm 0.0110}\)
        & \(\mathbf{0.7169 \pm 0.0105}\)
        \\

        \bottomrule
    \end{tabular}
    }
\caption{
Supporting \(K_{\mathrm{cand}}=1\) trajectory-geometry metrics at
\(6.4\,\mathrm{s}\) on the held-out test set.
Learned-model values are mean \(\pm\) sample standard deviation across
inference-noise seeds; CTRV is deterministic.
Lateral and longitudinal ADE are in metres; final heading error is in
degrees. Lower is better; bold indicates the lowest mean in each column.
}
\vspace{-1.0em}

\label{tab:geometry_results}
\end{table}

\subsection{Data Efficiency of Targeted Supervision}
\label{sec:results_data_efficiency}

\begin{figure}[t]
    \centering
    \includegraphics[width=0.98\columnwidth]
    {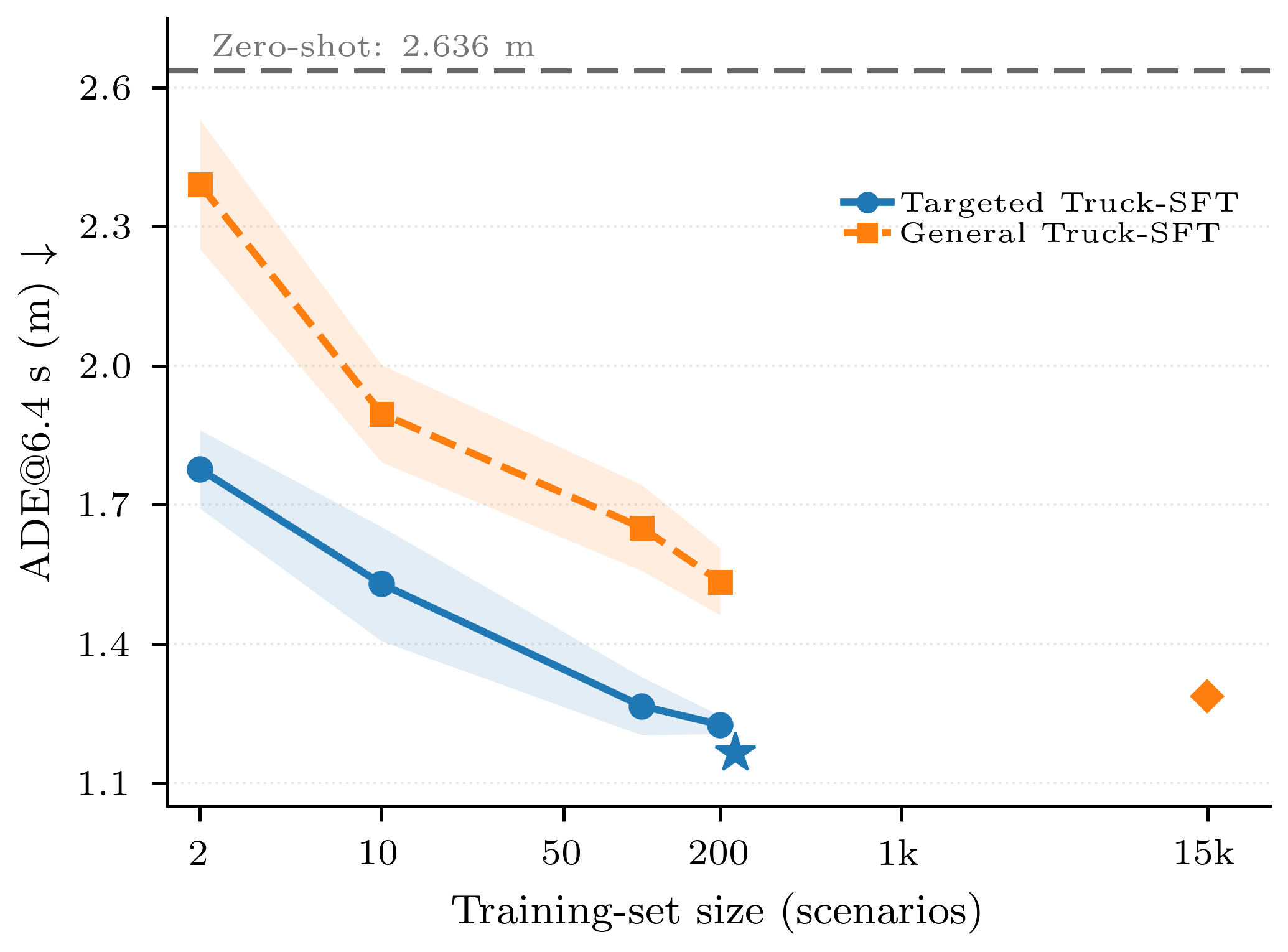}
\caption{
\textbf{Data efficiency of truck adaptation.}
ADE@\(6.4\,\mathrm{s}\) versus training-set size for targeted long-tail
and general truck-driving supervision.
Curves show mean performance across three independently trained subset
runs at each evaluated size; shaded bands denote \(\pm1\) standard
deviation.
The blue star and orange diamond denote the full-data Targeted Truck-SFT
(229 scenarios) and General Truck-SFT (14,891 scenarios) checkpoints,
respectively, shown as separate reference points.
The dashed gray line denotes the zero-shot Base model.
}
\vspace{-1.0em}

\label{fig:data_sweep}
\end{figure}

The gains obtained from 229 targeted scenarios motivate a closer
examination of data efficiency.
We ask whether targeted long-tail demonstrations provide more effective
supervision than the same number of general truck-driving scenarios.
Figure~\ref{fig:data_sweep} compares the two supervision sources at
matched training-scenario counts.
At the common budgets of 2, 10, 100, and 200 scenarios, targeted training
achieves approximately \(19\)-\(26\%\) lower mean
ADE@\(6.4\,\mathrm{s}\) than general truck-driving supervision on the
held-out long-tail distribution.
For targeted supervision, error decreases with training-set
size, with diminishing gains at larger evaluated budgets.
% The targeted curve also shows substantial gains with very limited data:
% training on two scenarios reduces ADE@\(6.4\,\mathrm{s}\) from the
% Base model's \(2.636\,\mathrm{m}\) to \(1.777\,\mathrm{m}\).
% Error further decreases to \(1.225\,\mathrm{m}\) at 200 scenarios,
% with diminishing gains over the evaluated budgets.

% TODO: Document subset selection and optimization settings for the
% sweep and full-data checkpoints.

% We next compare the full-data checkpoints
% (Table~\ref{tab:main_results}).
% For single-candidate prediction, General Truck-SFT is slightly more
% accurate at \(1\,\mathrm{s}\), whereas Targeted Truck-SFT achieves
% lower full-horizon error: \(1.165\) versus \(1.287\,\mathrm{m}\)
% ADE@\(6.4\,\mathrm{s}\), and \(3.159\) versus
% \(3.740\,\mathrm{m}\) FDE@\(6.4\,\mathrm{s}\).
% General$\rightarrow$Targeted SFT does not consistently improve over
% targeted-only adaptation.

We next compare the full-data checkpoints
(Table~\ref{tab:main_results}).
For single-candidate prediction, General Truck-SFT is slightly
more accurate at \(1\,\mathrm{s}\), whereas Targeted Truck-SFT
achieves lower full-horizon error: \(1.165\) versus
\(1.287\,\mathrm{m}\) ADE@\(6.4\,\mathrm{s}\), and
\(3.159\) versus \(3.740\,\mathrm{m}\) FDE@\(6.4\,\mathrm{s}\).
General$\rightarrow$Targeted SFT does not consistently
improve over targeted-only adaptation.
% The ordering differs under oracle best-of-16 evaluation:
% General Truck-SFT obtains the lowest minADE among the SFT variants,
% while General$\rightarrow$Targeted SFT obtains the lowest minFDE.
% Targeted supervision yields lower full-horizon ADE at matched scenario
% budgets, and a model trained on only 229 targeted scenarios remains
% competitive with one trained on approximately \(65\times\) as many
% general truck-driving scenarios.
The ordering differs under oracle best-of-16 evaluation:
General Truck-SFT obtains the lowest minADE among the SFT
variants, while General$\rightarrow$Targeted SFT obtains
the lowest minFDE.

Overall, Targeted Truck-SFT trained on only 229 targeted
scenarios remains competitive with General Truck-SFT trained
on approximately \(65\times\) as many general truck-driving
scenarios.

% \begin{figure*}[t]
%     \centering
%     \includegraphics[
%         height=0.9\textheight,
%         keepaspectratio
%     ]{figures/Exp1.png}
% \caption{
% \textbf{Illustrative adapt-then-steer results on held-out truck scenarios.}
% Columns I--IV progress chronologically, with Base
% Chain-of-Causation (CoC) annotations.
% Rows show Base, Targeted Truck-SFT, and Transformer FVS
% (top to bottom), using $K_{\mathrm{cand}}=1$ and a shared
% inference-noise seed.
% }
% \label{fig:qualitative_results}
% \end{figure*}

\begin{figure*}[p]
    \centering

    \ifsubmission
        % =========================
        % ICRA SUBMISSION VERSION
        % =========================
        \includegraphics[
            width=\textwidth,
            height=0.96\textheight,
            keepaspectratio
        ]{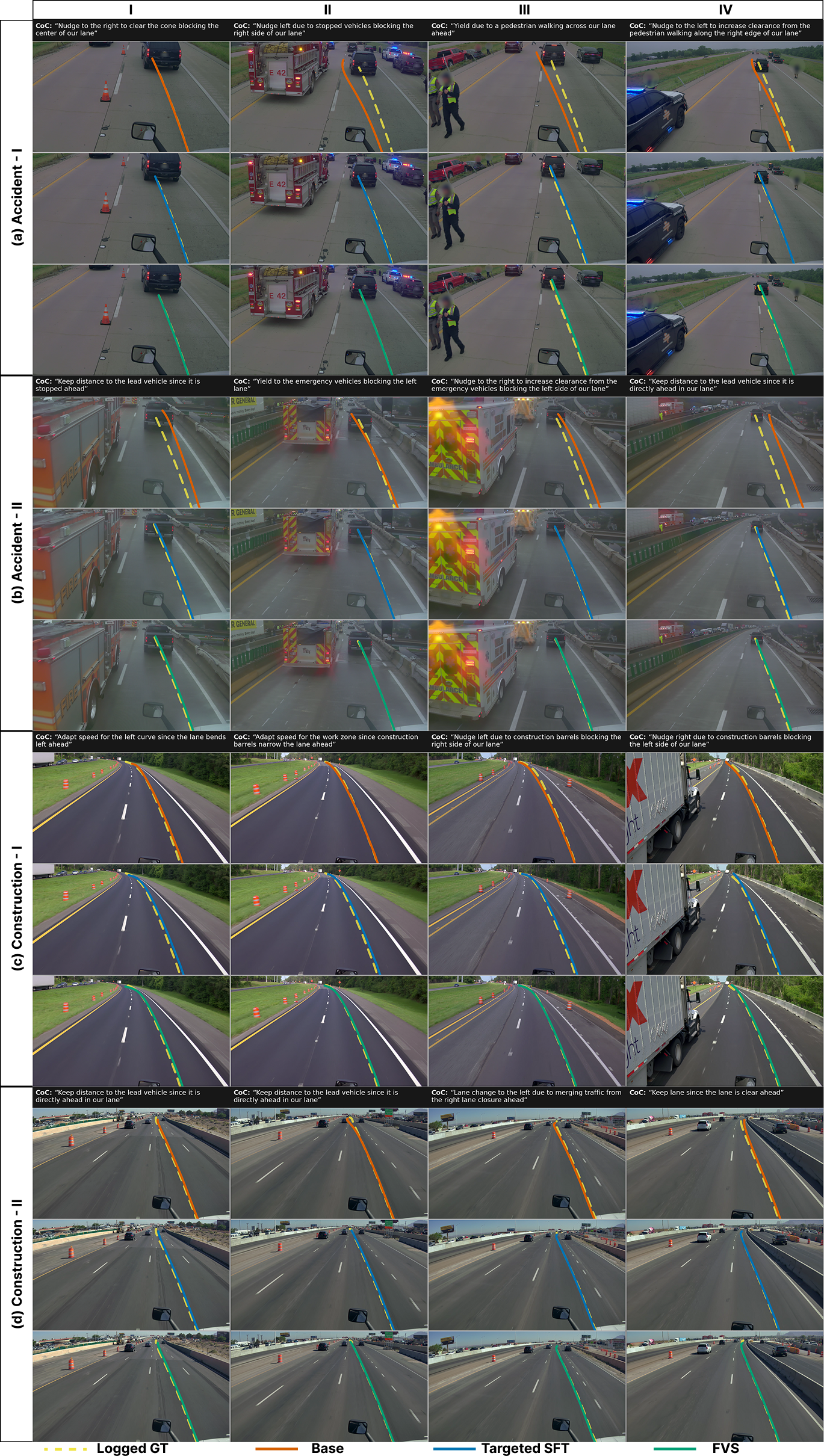}
         \caption{
        \textbf{Illustrative adapt-then-steer results on held-out truck scenarios.}
        Columns I--IV progress chronologically, with Base
        Chain-of-Causation (CoC) annotations.
        Rows show Base, Targeted Truck-SFT, and Transformer FVS
        (top to bottom), using $K_{\mathrm{cand}}=1$ and a shared
        inference-noise seed.
        }
    \else
        % =========================
        % ARXIV VERSION
        % =========================
        \includegraphics[
            width=\textwidth,
            height=0.96\textheight,
            keepaspectratio
        ]{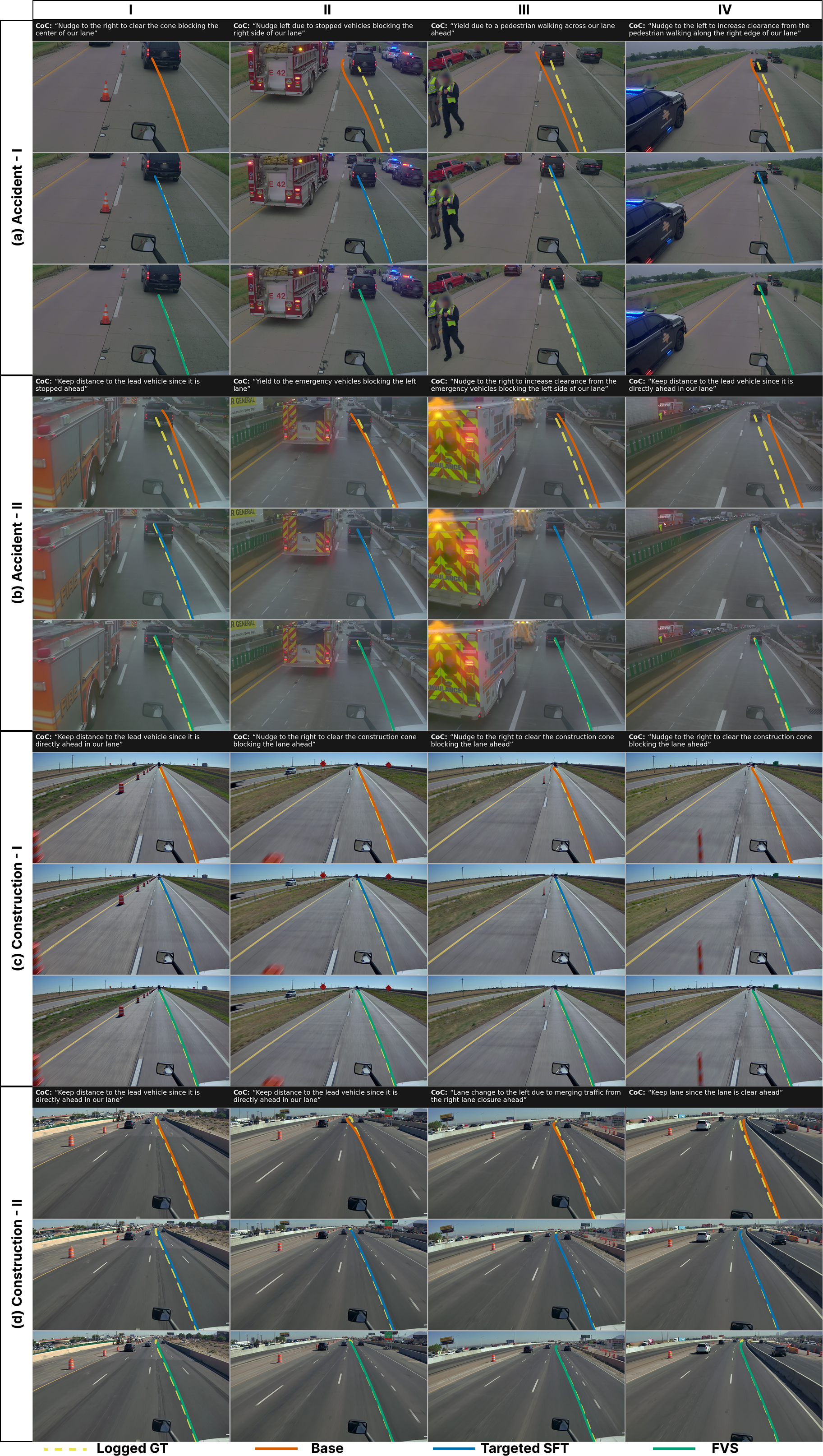}
         \caption{
        \textbf{Illustrative adapt-then-steer results on held-out truck scenarios.}
        Columns I--IV progress chronologically, with Base
        Chain-of-Causation (CoC) annotations.
        Rows compare Base, Targeted Truck-SFT, and Transformer FVS
    (top to bottom) against the logged truck trajectory (yellow dashed),
    using \(K_{\mathrm{cand}}=1\) and a shared inference-noise seed.
        }
    \fi

    % \caption{
    % \textbf{Illustrative adapt-then-steer results on held-out truck scenarios.}
    % Columns I--IV progress chronologically, with Base
    % Chain-of-Causation (CoC) annotations.
    % Rows show Base, Targeted Truck-SFT, and Transformer FVS
    % (top to bottom), using $K_{\mathrm{cand}}=1$ and a shared
    % inference-noise seed.
    % }
    \label{fig:qualitative_results}
\end{figure*}

% \begin{figure*}[p]
%     \centering
%     \includegraphics[
%         width=\textwidth,
%         height=0.96\textheight,
%         keepaspectratio
%     ]{figures/Fig4_red1.png}
%     \caption{
%     \textbf{Illustrative adapt-then-steer results on held-out truck scenarios.}
%     Columns I-IV progress chronologically, with Base
%     Chain-of-Causation (CoC) annotations.
%     Rows show Base, Targeted Truck-SFT, and Transformer FVS
%     (top to bottom), using $K_{\mathrm{cand}}=1$ and a shared
%     inference-noise seed.
%     }
%     \label{fig:qualitative_results}
% \end{figure*}

\subsection{Post-SFT Refinement with Flow Velocity Steering}
\label{sec:results_fvs}

% We next evaluate whether learned residuals can further reduce
% trajectory error using the same logged expert demonstrations while
% keeping Targeted Truck-SFT fixed.
% Table~\ref{tab:main_results} compares FVS (Sec.~\ref{sec:fvs}) with
% the final-action residual baseline.
% Relative to Targeted Truck-SFT, the Transformer final-action residual
% reduces full-horizon ADE and FDE by \(3.2\%\) and \(5.0\%\), respectively.
% Transformer FVS yields larger reductions of \(13.9\%\) and \(16.5\%\),
% with gains also extending to the shorter horizons.
% FVS outperforms the corresponding final-action residual for both
% MLP and Transformer architectures.

% We next evaluate whether FVS can further reduce trajectory
% error using the same logged expert demonstrations while
% keeping Targeted Truck-SFT fixed
% (Table~\ref{tab:main_results}).
We next evaluate whether FVS can further reduce trajectory error while keeping Targeted Truck-SFT fixed (Table~\ref{tab:main_results}), using the same 229 targeted training scenarios.
Relative to Targeted Truck-SFT, Transformer FVS reduces
full-horizon ADE and FDE by \(13.9\%\) and \(16.5\%\),
respectively, compared with \(3.2\%\) and \(5.0\%\) for
the Transformer final-action residual.
The FVS gains also extend to shorter horizons.
At matched trainable-parameter counts and training-update
budgets, FVS outperforms the corresponding final-action
residual for both MLP and Transformer architectures.

% The additional improvement extends across the held-out scenarios:
% Transformer FVS lowers scenario-mean ADE@\(6.4\,\mathrm{s}\)
% relative to Targeted Truck-SFT in all 26 scenarios.
% The mean percentage reductions are approximately \(16\%\) for
% accident-related scenarios and \(11\%\) for construction scenarios.
The additional FVS gain is likewise consistent across the
held-out set: Transformer FVS lowers scenario-mean
ADE@\(6.4\,\mathrm{s}\) relative to Targeted Truck-SFT in
all 26 scenarios, with mean reductions of approximately
\(16\%\) for accident-related scenarios and \(11\%\) for
construction scenarios.
% Transformer FVS also lowers all reported oracle minADE and minFDE
% values relative to Targeted Truck-SFT, improving both single-candidate
% and best-of-16 trajectory accuracy.
% Table~\ref{tab:geometry_results} additionally shows reductions in
% lateral ADE, longitudinal ADE, and final heading error, indicating
% improved lateral placement, longitudinal progress, and final orientation.
% Together, these comparisons support correction of the action-space
% flow velocity as an effective post-SFT refinement strategy relative
% to the evaluated final-action residuals.
% FVS adds one residual-network evaluation per existing Euler step,
% without additional flow-integration steps or vision-language
% backbone passes.
Transformer FVS also lowers all reported oracle minADE and minFDE
values relative to Targeted Truck-SFT, improving both single-candidate
and best-of-16 trajectory accuracy.
Table~\ref{tab:geometry_results} additionally shows reductions in
lateral ADE, longitudinal ADE, and final heading error, indicating
improved lateral placement, longitudinal progress, and final orientation.
Together, these comparisons support FVS as an effective post-SFT
refinement strategy.
% At inference, FVS adds one residual-network evaluation
% per existing Euler step, without additional flow-integration steps
% or vision-language backbone passes.
At inference, FVS requires one lightweight residual-network evaluation per existing Euler step, without increasing the number of flow-integration steps or vision-language backbone passes.

\subsection{Qualitative Analysis}
\label{sec:results_qualitative}

% Figure~\ref{fig:qualitative_results} shows selected illustrative
% timestamps from two accident-related and two construction scenarios,
% comparing Base, Targeted Truck-SFT, and Transformer FVS against the
% logged expert truck trajectories.
% In several displayed accident-related frames, Base's
% Chain-of-Causation describes a plausible high-level response, such as
% yielding or increasing clearance around emergency vehicles and
% pedestrians, while its generated trajectory differs substantially
% from the logged truck motion.
% Targeted Truck-SFT brings the predicted maneuver closer to the
% demonstrated expert trajectory.
% The Transformer FVS prediction either remains close to the SFT
% prediction or shows a smaller correction toward the logged motion.
% The construction examples show a similar pattern under curved
% work-zone geometry, temporary lane constraints, and adjacent-truck
% interactions.
% These examples are consistent with the aggregate quantitative results:
% targeted SFT provides the primary trajectory correction, while FVS
% further refines the adapted predictions.
% Complete temporal sequences are provided in the supplementary videos.

% \subsection{Qualitative Analysis}
% \label{sec:results_qualitative}

\ifsubmission
% =========================================================
% ICRA SUBMISSION VERSION
% =========================================================

Figure~\ref{fig:qualitative_results} shows selected illustrative
timestamps from two accident-related and two construction scenarios,
comparing Base, Targeted Truck-SFT, and Transformer FVS against the
logged expert truck trajectories.
In several displayed accident-related frames, Base's
Chain-of-Causation describes a plausible high-level response, such as
yielding or increasing clearance around emergency vehicles and
pedestrians, while its generated trajectory differs substantially
from the logged truck motion.
Targeted Truck-SFT brings the predicted maneuver closer to the
demonstrated expert trajectory.
The Transformer FVS prediction either remains close to the SFT
prediction or shows a smaller correction toward the logged motion.
The construction examples show a similar pattern under curved
work-zone geometry, temporary lane constraints, and adjacent-truck
interactions.
These examples are consistent with the aggregate quantitative results:
targeted SFT provides the primary trajectory correction, while FVS
further refines the adapted predictions.
Complete temporal sequences are provided in the supplementary videos.

\else
% =========================================================
% ARXIV VERSION
% =========================================================

% Put your revised/expanded arXiv qualitative-analysis text here.

Figure~\ref{fig:qualitative_results} shows selected illustrative
timestamps from two accident-related and two construction scenarios,
comparing Base, Targeted Truck-SFT, and Transformer FVS against the
logged expert truck trajectories.
In several displayed accident-related frames, Base's
Chain-of-Causation describes a plausible high-level response, such as
yielding or increasing clearance around emergency vehicles and
pedestrians, while its generated trajectory differs substantially
from the logged truck motion.
Targeted Truck-SFT brings the predicted maneuver closer to the
demonstrated expert trajectory.
The Transformer FVS prediction either remains close to the SFT
prediction or shows a smaller correction toward the logged motion.
The construction examples show a similar pattern under temporary
work-zone constraints, including cone intrusions and lane-closure
maneuvers.
These examples are consistent with the aggregate quantitative results:
targeted SFT provides the primary trajectory correction, while FVS
further refines the adapted predictions.
Complete temporal sequences are provided in the supplementary videos.

\fi
%==============================================================================
% \vspace{-3pt}
\section{Conclusion}
\label{sec:conclusion} 
We studied whether an autonomous-driving VLA can be adapted to Class~8 trucks without retraining its vision-language backbone. With the backbone frozen, fine-tuning the action-generation stack on 229 scenario-aligned truck events reduces full-horizon ADE and FDE by approximately \(56\%\).
% and remains competitive with General Truck-SFT 
% trained on approximately \(65\times\). 
% as many general truck-driving scenarios;
% trained on approximately \(65\times\) more general truck-driving data; 
% matched-budget comparisons consistently favor scenario-aligned supervision. Starting from this adapted policy, 
At matched data budgets, scenario-aligned supervision consistently yields lower full-horizon ADE than general truck-driving supervision, while Targeted Truck-SFT remains
competitive with General Truck-SFT trained on approximately \(65\times\) as
many general truck-driving scenarios. Starting from Targeted Truck-SFT, 
Flow Velocity Steering (FVS) further reduces full-horizon ADE and FDE by \(13.9\%\) and \(16.5\%\), respectively, while keeping the adapted VLA fixed. 
% Across both residual architectures, steering the action-space velocity field throughout generation outperforms the corresponding final-action correction. 
Together, these results support an \emph{adapt-then-steer} strategy for vehicle-domain transfer to Class~8 trucks: preserve the pretrained vision-language backbone, specialize action generation with scenario-aligned demonstrations, and refine trajectory generation
by steering the adapted policy's action-space flow velocity.
Our evaluation measures open-loop agreement with logged truck
trajectories rather than closed-loop driving performance.
Future work will extend this study to closed-loop evaluation and simulator-based post-training in a validated vision-based truck simulator with realistic vehicle dynamics.

\vspace{-0.3em}

\bibliographystyle{IEEEtran}
\bibliography{citations.bib}

\end{document}